\documentclass[11pt]{article}

\usepackage{hyperref}
\usepackage{fullpage}
\usepackage{amsmath,amsfonts,amsthm,amssymb,xspace,bm,verbatim,dsfont,mathtools}
\usepackage{graphicx}
\usepackage{url,algorithm,algorithmicx}
\usepackage{algpseudocode}
\usepackage{color}
\usepackage{epstopdf}
\usepackage[titletoc,title]{appendix}
\usepackage{comment}

\usepackage{graphicx,caption,subcaption}
\usepackage[numbers]{natbib}

\theoremstyle{plain}
\newtheorem{theorem}{Theorem}[section]
\newtheorem{lemma}[theorem]{Lemma}
\newtheorem*{lemma*}{Lemma}

\theoremstyle{definition}

\newcommand{\note}[1]{\marginpar{\tiny *note in TeX*}}
\newcommand{\ignore}[1]{}

\renewcommand{\phi}{\varphi}

\newcommand{\aff}{{\mathcal A}}

\begin{document}
\title{Greedy Subspace Clustering}
\author{
Dohyung Park\\
{The University of Texas at Austin}\\
{\href{mailto:dhpark@utexas.edu}{dhpark@utexas.edu}}
\and
Constantine Caramanis\\
{The University of Texas at Austin}\\
{\href{mailto:constantine@utexas.edu}{constantine@utexas.edu}}
\and
Sujay Sanghavi\\
{The University of Texas at Austin}\\
{\href{mailto:sanghavi@mail.utexas.edu}{sanghavi@mail.utexas.edu}}
}
\maketitle

\def\aff{\mathrm{aff}}
\def\sp{\mathrm{span}}
\def\range{\mathrm{Range}}
\def\proj{\mathrm{Proj}}
\def\set{\mathcal}

\newtheorem{proposition}[theorem]{Proposition} 
\newtheorem{fact}[theorem]{Fact} 
\newtheorem{remark}[theorem]{Remark}


\begin{abstract} 
We consider the problem of subspace clustering: given points that lie on or near the {\em union} of many low-dimensional linear subspaces, recover the subspaces. To this end, one first identifies sets of points close to the same subspace and uses the sets to estimate the subspaces. As the geometric structure of the clusters (linear subspaces) forbids proper performance of general distance based approaches such as $K$-means, many model-specific methods have been proposed. In this paper, we provide new simple and efficient algorithms for this problem. Our statistical analysis shows that the algorithms are guaranteed exact (perfect) clustering performance under certain conditions on the number of points and the affinity between subspaces. These conditions are weaker than those considered in the standard statistical literature. Experimental results on synthetic data generated from the standard unions of subspaces model demonstrate our theory. We also show that our algorithm performs competitively against state-of-the-art algorithms on real-world applications such as motion segmentation and face clustering, with much simpler implementation and lower computational cost.
\end{abstract}

\section{Introduction}

Subspace clustering is a classic problem where one is given points in a high-dimensional ambient space and would like to approximate them by a {\em union} of lower-dimensional linear subspaces. In particular, each subspace contains a subset of the points. This problem is hard because one needs to jointly find the subspaces, and the points corresponding to each; the data we are given are unlabeled. The unions of subspaces model naturally arises in settings where data from multiple latent phenomena are mixed together and need to be separated. Applications of subspace clustering include motion segmentation \cite{vidal2008multiframe}, face clustering \cite{ho2003clustering}, gene expression analysis \cite{kriegel2009clustering}, and system identification \cite{vidal2003algebraic
}. In these applications, data points with the same label (e.g., face images of a person under varying illumination conditions, feature points of a moving rigid object in a video sequence) lie on a low-dimensional subspace, and the mixed dataset can be modeled by unions of subspaces. For detailed description of the applications, we refer the readers to the reviews \cite{kriegel2009clustering, vidal2011subspace} and references therein.

There is now a sizable literature on empirical methods for this particular problem and some statistical analysis as well. Many recently proposed methods, which perform remarkably well and have theoretical guarantees on their performances, can be characterized as involving two steps: {\em (a)} finding a ``neighborhood" for each data point, and {\em (b)} finding the subspaces and/or clustering the points given these neighborhoods. Here, neighbors of a point are other points that the algorithm estimates to lie on the same subspace as the point (and not necessarily just closest in Euclidean distance).

{\bf Our contributions:} In this paper we devise {\bf new algorithms} for each of the two steps above; {\em (a)} we develop a new method, Nearest Subspace Neighbor (NSN), to determine a neighborhood set for each point, and {\em (b)} a new method, Greedy Subspace Recovery (GSR), to recover subspaces from given neighborhoods. Each of these two methods can be used in conjunction with other methods for the corresponding other step; however, in this paper we focus on two algorithms that use NSN followed by GSR and Spectral clustering, respectively. Our main result is establishing {\bf statistical guarantees for exact clustering with general subspace conditions}, in the standard models considered in recent analytical literature on subspace clustering. Our condition for exact recovery is weaker than the conditions of other existing algorithms that only guarantee \emph{correct neighborhoods}\footnote{By correct neighborhood, we mean that for each point every neighbor point lies on the same subspace.}, which do not always lead to correct clustering. We provide numerical results which demonstrate our theory. We also show that for the real-world applications our algorithm performs competitively against those of state-of-the-art algorithms, but the computational cost is much lower than them. Moreover, our algorithms are much simpler to implement.

\begin{table*}[t]
\begin{center}
\begin{scriptsize}
\begin{tabular}{c||c|c|cc}
	\hline
	 & & Subspace     & Conditions for: & 		      \\
	Algorithm & What is guaranteed & condition    & Fully random model & Semi-random model    \\
	\hline
	SSC \citep{elhamifar2013TPAMI, soltanolkotabi2012geometric} & Correct neighborhoods & None & $\frac{d}{p} = O(\frac{\log(n/d)}{\log(nL)})$ & $\max\aff = O(\frac{\sqrt{\log(n/d)}}{\log(nL)})$ \\
	LRR \citep{liu2013TPAMI} & Exact clustering & No intersection & - & - \\
	SSC-OMP \citep{dyer2013greedy} & Correct neighborhoods & No intersection & - & - \\
	TSC \citep{heckel2013thresholding,heckel2014robust} & Exact clustering & None & $\frac{d}{p} = O(\frac{1}{\log(nL)})$ & $\max\aff = O(\frac{1}{\log(nL)})$ \\
	LRSSC \citep{wang2013nips} & Correct neighborhoods & None & $\frac{d}{p} = O(\frac{1}{\log(nL)})$ & - \\
	\hline
	NSN+GSR & Exact clustering & None & $\frac{d}{p} = O(\frac{\log n}{\log (ndL)})$ & $\max\aff = O(\sqrt{\frac{\log n}{(\log dL) \cdot \log(ndL)}})$ \\
	NSN+Spectral & Exact clustering & None & $\frac{d}{p} = O(\frac{\log n}{\log (ndL)})$ & - \\
	\hline
\end{tabular}
\end{scriptsize}
\end{center}
\caption{Subspace clustering algorithms with theoretical guarantees. LRR and SSC-OMP have only deterministic guarantees, not statistical ones. In the two standard statistical models, there are $n$ data points on each of $L$ $d$-dimensional subspaces in $\mathbb{R}^p$. For the definition of $\max\aff$, we refer the readers to Section \ref{sec:model}.}
\label{tab:algorithms}
\end{table*}

\subsection{Related work}
The problem was first formulated in the data mining community 
\cite{kriegel2009clustering}. Most of the related work in this field assumes that an underlying subspace is parallel to some canonical axes. Subspace clustering for unions of arbitrary subspaces is considered mostly in the machine learning and the computer vision communities \citep{vidal2011subspace}. Most of the results from those communities are based on empirical justification. They provided algorithms derived from theoretical intuition and showed that they perform empirically well with practical dataset. To name a few, GPCA \cite{vidal2003gpca
}, Spectral curvature clustering (SCC) \cite{chen2009spectral}, and many iterative methods \cite{bradley2000k, tseng2000kmeans, zhang2012hybrid} show their good empirical performance for subspace clustering. However, they lack theoretical analysis that guarantees exact clustering.

As described above, several algorithms with a common structure are recently proposed with both theoretical guarantees and remarkable empirical performance. \citet{elhamifar2013TPAMI} proposed an algorithm called Sparse Subspace Clustering (SSC), which uses $\ell_1$-minimization for neighborhood construction. They proved that if the subspaces have \emph{no intersection}\footnote{By no intersection between subspaces, we mean that they share only the null point.}, SSC always finds a correct neighborhood matrix. Later, \citet{soltanolkotabi2012geometric} provided a statistical guarantee of the algorithm for subspaces with intersection. \citet{dyer2013greedy} proposed another algorithm called SSC-OMP, which uses Orthogonal Matching Pursuit (OMP) instead of $\ell_1$-minimization in SSC. Another algorithm called Low-Rank Representation (LRR) which uses nuclear norm minimization is proposed by \citet{liu2013TPAMI}. \citet{wang2013nips} proposed an hybrid algorithm, Low-Rank and Sparse Subspace Clustering (LRSSC), which involves both $\ell_1$-norm and nuclear norm. \citet{heckel2013thresholding} presented Thresholding based Subspace Clustering (TSC), which constructs neighborhoods based on the inner products between data points. All of these algorithms use spectral clustering for the clustering step. 

The analysis in those papers focuses on neither exact recovery of the subspaces nor exact clustering in general subspace conditions. SSC, SSC-OMP, and LRSSC only guarantee correct neighborhoods which do not always lead to exact clustering. LRR guarantees exact clustering only when the subspaces have no intersections. In this paper, we provide novel algorithms that guarantee exact clustering in general subspace conditions. When we were preparing this manuscript, it is proved that TSC guarantees exact clustering under certain conditions \citep{heckel2014robust}, but the conditions are stricter than ours. (See Table \ref{tab:algorithms})

\subsection{Notation}
There is a set of $N$ data points in $\mathbb{R}^p$, denoted by $\set{Y} = \{y_1, \ldots, y_N\}$. The data points are lying on or near a union of $L$ subspaces $\set{D} = \cup_{i=1}^L \set{D}_i$. Each subspace $\set{D}_i$ is of dimension $d_i$ which is smaller than $p$. For each point $y_j$, $w_j$ denotes the index of the nearest subspace. Let $N_i$ denote the number of points whose nearest subspace is $\set{D}_i$, i.e., $N_i = \sum_{j=1}^N \mathbb{I}_{w_j = i}$. Throughout this paper, sets and subspaces are denoted by calligraphic letters. Matrices and key parameters are denoted by letters in upper case, and vectors and scalars are denoted by letters in lower case. We frequently denote the set of $n$ indices by $[n] = \{1,2,\ldots,n\}$. As usual, $\sp\{\cdot\}$ denotes a subspace spanned by a set of vectors. For example, $\sp\{v_1,\ldots,v_n\} = \{v : v = \sum_{i=1}^n \alpha_i v_i, \alpha_1,\ldots,\alpha_n \in \mathbb{R}\}$. $\proj_{\set{U}} y$ is defined as the projection of $y$ onto subspace $\set{U}$. That is, $\proj_{\set{U}} y = \arg \min_{u \in \set{U}} \|y-u\|_2$. $\mathbb{I}\{\cdot\}$ denotes the indicator function which is one if the statement is true and zero otherwise. Finally, $\bigoplus$ denotes the direct sum.


\renewcommand{\algorithmicrequire}{\textbf{Input:}}
\renewcommand{\algorithmicensure}{\textbf{Output:}}	

\section{Algorithms}

We propose two algorithms for subspace clustering as follows.
\begin{itemize}
	\item NSN+GSR : Run Nearest Subspace Neighbor (NSN) to construct a neighborhood matrix $W \in \{0,1\}^{N \times N}$, and then run Greedy Subspace Recovery (GSR) for $W$.
	\item NSN+Spectral : Run Nearest Subspace Neighbor (NSN) to construct a neighborhood matrix $W \in \{0,1\}^{N \times N}$, and then run spectral clustering for $Z = W + W^\top$.
\end{itemize}	

\subsection{Nearest Subspace Neighbor (NSN)}

NSN approaches the problem of \emph{finding neighbor points most likely to be on the same subspace} in a greedy fashion. At first, given a point $y$ without any other knowledge, the one single point that is most likely to be a neighbor of $y$ is the nearest point of the line $\sp\{y\}$. In the following steps, if we have found a few correct neighbor points (lying on the same true subspace) and have no other knowledge about the true subspace and the rest of the points, then the most potentially correct point is the one closest to the subspace spanned by the correct neighbors we have. This motivates us to propose NSN described in the following.

\begin{algorithm}
	\caption{Nearest Subspace Neighbor (NSN)}
\begin{algorithmic} \label{alg:nsn}
	\Require A set of $N$ samples $\mathcal{Y} = \{y_1,\ldots,y_N\}$, The number of required neighbors $K$, Maximum subspace dimension $k_{\max}$.
	\Ensure A neighborhood matrix $W \in \{0,1\}^{N \times N}$
	\State $y_i \gets y_i / \|y_i\|_2$, $\forall i \in [N]$ \Comment{Normalize magnitudes}
	\For {$i = 1,\ldots,N$} \Comment{Run NSN for each data point}
		\State $\set{I}_i \gets \{i\}$
		\For {$k = 1,\ldots,K$} \Comment{Iteratively add the closest point to the current subspace}
			\If {$k \le k_{\max}$}
				\State {$\set{U} \gets \sp\{y_j:j \in \set{I}_i\}$}
			\EndIf
			\State $j^* \gets \arg\max_{j \in [N] \setminus \set{I}_i} \|\proj_{\set{U}} y_j\|_2$
			\State $\set{I}_i \gets \set{I}_i \cup \{j^*\}$
		\EndFor
		\State $W_{ij} \gets \mathbb{I}_{j \in \set{I}_i \text{ or } y_j \in \set{U}} ,~ \forall j \in [N]$ \Comment{Construct the neighborhood matrix}
	\EndFor
\end{algorithmic}
\end{algorithm}

NSN collects $K$ neighbors sequentially for each point. At each step $k$, a $k$-dimensional subspace $\set{U}$ spanned by the point and its $k-1$ neighbors is constructed, and the point closest to the subspace is newly collected. After $k \ge k_{\max}$, the subspace $\set{U}$ constructed at the $k_{\max}$th step is used for collecting neighbors. At last, if there are more points lying on $\set{U}$, they are also counted as neighbors. The subspace $\set{U}$ can be stored in the form of a matrix $U \in \mathbb{R}^{p \times \text{dim}(\set{U})}$ whose columns form an orthonormal basis of $\set{U}$. Then $\|\proj_{\set{U}} y_j\|_2$ can be computed easily because it is equal to $\|U^\top y_j\|_2$. While a naive implementation requires $O(K^2pN^2)$ computational cost, this can be reduced to $O(KpN^2)$, and the faster implementation is described in Section \ref{sec:fastimpl}. We note that this computational cost is much lower than that of the convex optimization based methods (e.g., SSC \cite{elhamifar2013TPAMI} and LRR \cite{liu2013TPAMI}) which solve a convex program with $N^2$ variables and $pN$ constraints.


NSN for subspace clustering shares the same philosophy with Orthogonal Matching Pursuit (OMP) for sparse recovery in the sense that it incrementally picks the point (dictionary element) that is the most likely to be correct, assuming that the algorithms have found the correct ones.
In subspace clustering, that point is the one closest to the subspace spanned by the currently selected points, while in sparse recovery it is the one closest to the residual of linear regression by the selected points. In the sparse recovery literature, the performance of OMP is shown to be comparable to that of Basis Pursuit ($\ell_1$-minimization) both theoretically and empirically \citep{tropp2007signal, kunis2008random}. One of the contributions of this work is to show that this high-level intuition is indeed born out, provable, as we show that NSN also performs well in collecting neighbors lying on the same subspace.

\subsection{Greedy Subspace Recovery (GSR)}

Suppose that NSN has found correct neighbors for a data point. How can we check if they are indeed correct, that is, lying on the same true subspace? One natural way is to count the number of points close to the subspace spanned by the neighbors. If they span one of the true subspaces, then many other points will be lying on the span. If they do not span any true subspaces, few points will be close to it. This fact motivates us to use a greedy algorithm to recover the subspaces. Using the neighborhood constructed by NSN (or some other algorithm), we recover the $L$ subspaces. If there is a neighborhood set containing only the points on the same subspace for each subspace, the algorithm successfully recovers the unions of the true subspaces exactly.

\begin{algorithm}
	\caption{Greedy Subspace Recovery (GSR)}
\begin{algorithmic} \label{alg:gsr}
	\Require $N$ points $\set{Y} = \{y_1,\ldots,y_N\}$, A neighborhood matrix $W \in \{0,1\}^{N \times N}$, Error bound $\epsilon$
	\Ensure Estimated subspaces $\hat{\set{D}} = \cup_{l=1}^L \hat{D}_l$. Estimated labels $\hat{w}_1, \ldots, \hat{w}_N$

	\State $y_i \gets y_i / \|y_i\|_2$, $\forall i \in [N]$ \Comment{Normalize magnitudes}
	\State $\set{W}_i \gets \text{Top-$d$} \{ y_j : W_{ij} = 1 \} $, $\forall i \in [N]$ \Comment{Estimate a subspace using the neighbors for each point}
	
	\State $\set{I} \gets [N]$
	\While {$\set{I} \neq \emptyset$}\Comment{Iteratively pick the best subspace estimates}
		\State $i^* \gets \arg \max_{i \in \set{I}} \sum_{j=1}^N \mathbb{I}\{ \|\proj_{\set{W}_i} y_j\|_2 \ge 1 - \epsilon \}$
		\State $\hat{\set{D}}_l \gets \hat{\set{W}}_{i^*}$
		\State $\set{I} \gets \set{I} \setminus \{j : \|\proj_{\set{W}_{i^*}} y_j\|_2 \ge 1 - \epsilon \}$
	\EndWhile	
	
	\State $\hat{w}_i \gets \arg\max_{l \in [L]} \|\proj_{\hat{D}_l} y_i\|_2,~ \forall i \in [N]$ \Comment{Label the points using the subspace estimates}
\end{algorithmic}
\end{algorithm}

Recall that the matrix $W$ contains the labelings of the points, so that $W_{ij} = 1$ if point $i$ is assigned to subspace $j$. Top-$d \{ y_j : W_{ij} = 1 \}$ denotes the $d$-dimensional principal subspace of the set of vectors $\{ y_j : W_{ij} = 1 \}$. This can be obtained by taking the first $d$ left singular vectors of the matrix whose columns are the vector in the set. If there are only $d$ vectors in the set, Gram-Schmidt orthogonalization will give us the subspace. As in NSN, it is efficient to store a subspace $\set{W}_i$ in the form of its orthogonal basis because we can easily compute the norm of a projection onto the subspace.

Testing a candidate subspace by counting the number of near points has already been considered in the subspace clustering literature. In \citep{yang2006robust}, the authors proposed to run RANdom SAmple Consensus (RANSAC) iteratively. RANSAC randomly selects a few points and checks if there are many other points near the subspace spanned by the collected points. Instead of randomly choosing sample points, GSR receives some candidate subspaces (in the form of sets of points) from NSN (or possibly some other algorithm) and selects subspaces in a greedy way as specified in the algorithm above.

\section{Theoretical results} \label{sec:theory}

We analyze our algorithms in two standard noiseless models. The main theorems present sufficient conditions under which the algorithms cluster the points exactly with high probability. For simplicity of analysis, we assume that every subspace is of the same dimension, and the number of data points on each subspace is the same, i.e.,
$d \triangleq d_1 = \cdots = d_L ,\quad n \triangleq N_1 = \cdots = N_L$. We assume that $d$ is known to the algorithm. Nonetheless, our analysis can extend to the general case. 

\subsection{Statistical models} \label{sec:model}

We consider two models which have been used in the subspace clustering literature:
\begin{itemize}
\item Fully random model: The subspaces are drawn iid uniformly at random, and the points are also iid randomly generated.
\item Semi-random model: The subspaces are arbitrarily determined, but the points are iid randomly generated.
\end{itemize}

Let $D_i \in \mathbb{R}^{p \times d}, i \in [L]$ be a matrix whose columns form an orthonormal basis of $\set{D}_i$.
An important measure that we use in the analysis is the {\em affinity} between two subspaces, defined as
\begin{align*}
\aff(i,j) \triangleq \frac{\|D_i^\top D_j\|_F}{\sqrt{d}} = \sqrt{\frac{\sum_{k=1}^{d} \cos^2 \theta^{i,j}_k}{d}} \in [0,1],
\end{align*}
where $\theta^{i,j}_k$ is the $k$th principal angle between $\set{D}_i$ and $\set{D}_j$. Two subspaces $\set{D}_i$ and $\set{D}_j$ are identical if and only if $\aff(i,j) = 1$. If $\aff(i,j) = 0$, every vector on $\set{D}_i$ is orthogonal to any vectors on $\set{D}_j$. We also define the maximum affinity as
$$
\max \aff \triangleq \max_{i,j \in [L], i \neq j} \aff(i,j) \in [0,1].
$$

There are $N = nL$ points, and there are $n$ points exactly lying on each subspace. We assume that each data point $y_i$ is drawn iid uniformly at random from $\mathbb{S}^{p-1} \cap \set{D}_{w_i}$ where $\mathbb{S}^{p-1}$ is the unit sphere in $\mathbb{R}^p$. Equivalently,
\begin{align*}
y_i = D_{w_i} x_i ,\quad x_i \sim \text{Unif}(\mathbb{S}^{d-1}) ,\quad \forall i \in [N].
\end{align*}

As the points are generated randomly on their corresponding subspaces, there are no points lying on an intersection of two subspaces, almost surely. This implies that \emph{with probability one the points are clustered correctly provided that the true subspaces are recovered exactly}.

\subsection{Main theorems}
The first theorem gives a statistical guarantee for the fully random model.
\begin{theorem} \label{thm:NSNfullyrandom}
Suppose $L$ $d$-dimensional subspaces and $n$ points on each subspace are generated in the fully random model with $n$ polynomial in $d$. There are constants $C_1, C_2 > 0$ such that if
\begin{align} \label{eqn:NSNfullyrandom}
\frac{n}{d} > C_1 \left( \log \frac{ne}{d\delta} \right)^2 ,\quad \frac{d}{p} < \frac{C_2 \log n}{\log (ndL \delta^{-1})},
\end{align}
then with probability at least $1 - \frac{3L\delta}{1-\delta}$, NSN+GSR\footnote{NSN with $K = k_{max} = d$ followed by GSR with arbitrarily small $\epsilon$.} clusters the points exactly. Also, there are other constants $C_1', C_2' > 0$ such that if \eqref{eqn:NSNfullyrandom} with $C_1$ and $C_2$ replaced by $C_1'$ and $C_2'$ holds then NSN+Spectral\footnote{NSN with $K = k_{max} = d$.} clusters the points exactly with probability at least $1 - \frac{3L\delta}{1-\delta}$. $e$ is the exponential constant.
\end{theorem}

Our sufficient conditions for exact clustering explain when subspace clustering becomes easy or difficult, and they are consistent with our intuition. For NSN to find correct neighbors, the points on the same subspace should be many enough so that they look like lying on a subspace. This condition is spelled out in the first inequality of \eqref{eqn:NSNfullyrandom}. We note that the condition holds even when $n/d$ is a constant, i.e., $n$ is linear in $d$. The second inequality implies that the dimension of the subspaces should not be too high for subspaces to be distinguishable. If $d$ is high, the random subspaces are more likely to be close to each other, and hence they become more difficult to be distinguished. However, as $n$ increases, the points become dense on the subspaces, and hence it becomes easier to identify different subspaces.

Let us compare our result with the conditions required for success in the fully random model in the existing literature. In \cite{soltanolkotabi2012geometric}, it is required for SSC to have correct neighborhoods that $n$ should be superlinear in $d$ when $d/p$ fixed. In \cite{heckel2013thresholding, wang2013nips}, the conditions on $d/p$ becomes worse as we have more points. On the other hand, our algorithms are guaranteed exact clustering of the points, and the sufficient condition is order-wise at least as good as the conditions for correct neighborhoods by the existing algorithms (See Table \ref{tab:algorithms}). Moreover, exact clustering is guaranteed even when $n$ is linear in $d$, and $d/p$ fixed.

For the semi-random model, we have the following general theorem.
\begin{theorem} \label{thm:NSNsemirandom}
Suppose $L$ $d$-dimensional subspaces are arbitrarily chosen, and $n$ points on each subspace are generated in the semi-random model with $n$ polynomial in $d$. There are constants $C_1, C_2 > 0$ such that if
\begin{align}
\frac{n}{d} > C_1 \left( \log \frac{ne}{d\delta} \right)^2 ,\quad \max\aff < \sqrt{\frac{C_2 \log n}{\log (dL\delta^{-1}) \cdot \log (ndL\delta^{-1})}}. \label{eqn:NSNsemirandom}
\end{align}
then with probability at least $1 - \frac{3L\delta}{1-\delta}$, NSN+GSR\footnote{NSN with $K = d-1$ and $k_{max} = \lceil 2 \log d \rceil$ followed by GSR with arbitrarily small $\epsilon$.} clusters the points exactly.
\end{theorem}

In the semi-random model, the sufficient condition does not depend on the ambient dimension $p$. When the affinities between subspaces are fixed, and the points are exactly lying on the subspaces, the difficulty of the problem does not depend on the ambient dimension. It rather depends on $\max\aff$, which measures how close the subspaces are. As they become closer to each other, it becomes more difficult to distinguish the subspaces. The second inequality of \eqref{eqn:NSNsemirandom} explains this intuition. The inequality also shows that if we have more data points, the problem becomes easier to identify different subspaces. 

Compared with other algorithms, NSN+GSR is guaranteed exact clustering, and more importantly, the condition on $\max\aff$ improves as $n$ grows. This remark is consistent with the practical performance of the algorithm which improves as the number of data points increases, while the existing guarantees of other algorithms are not. In \cite{soltanolkotabi2012geometric}, correct neighborhoods in SSC are guaranteed if $\max\aff = O(\sqrt{\log (n/d)}/\log(nL))$. In \cite{heckel2013thresholding}, exact clustering of TSC is guaranteed if $\max\aff = O(1/\log(nL))$. However, these algorithms perform empirically better as the number of data points increases.


\section{Experimental results} \label{sec:empirical}

In this section, we empirically compare our algorithms with the existing algorithms in terms of clustering performance and computational time (on a single desktop). For NSN, we used the fast implementation described in Section \ref{sec:fastimpl}. The compared algorithms are $K$-means, $K$-flats\footnote{$K$-flats is similar to $K$-means. At each iteration, it computes top-$d$ principal subspaces of the points with the same label, and then labels every point based on its distances to those subspaces.}, SSC, LRR, SCC, TSC\footnote{The MATLAB codes for SSC, LRR, SCC, and TSC are obtained from \url{http://www.cis.jhu.edu/~ehsan/code.htm}, \url{https://sites.google.com/site/guangcanliu}, and \url{http://www.math.duke.edu/~glchen/scc.html}, \url{http://www.nari.ee.ethz.ch/commth/research/downloads/sc.html}, respectively.}, and SSC-OMP\footnote{For each data point, OMP constructs a neighborhood for each point by regressing the point on the other points up to $10^{-4}$ accuracy.}. The numbers of replicates in $K$-means, $K$-flats, and the $K$-means used in the spectral clustering are all fixed to $10$. The algorithms are compared in terms of \emph{Clustering error (CE)} and \emph{Neighborhood selection error (NSE)}, defined as
\begin{align*}
(\text{CE}) = \min_{\pi \in \Pi_L} \frac{1}{N} \sum_{i=1}^N \mathbb{I}(w_i \neq \pi(\hat{w}_i)) ,\quad
(\text{NSE}) = \frac{1}{N} \sum_{i=1}^N \mathbb{I}(\exists j : W_{ij} \neq 0, w_i \neq w_j)
\end{align*}
where $\Pi_L$ is the permutation space of $[L]$. CE is the proportion of incorrectly labeled data points. Since clustering is invariant up to permutation of label indices, the error is equal to the minimum disagreement over the permutation of label indices. 
NSE measures the proportion of the points which do not have all correct neighbors.
\footnote{For the neighborhood matrices from SSC, LRR, and SSC-OMP, the $d$ points with the maximum weights are regarded as neighbors for each point. For TSC, the $d$ nearest neighbors are collected for each point.}

\subsection{Synthetic data}

\begin{figure}[t]
\begin{center}
	\includegraphics[width=\textwidth]{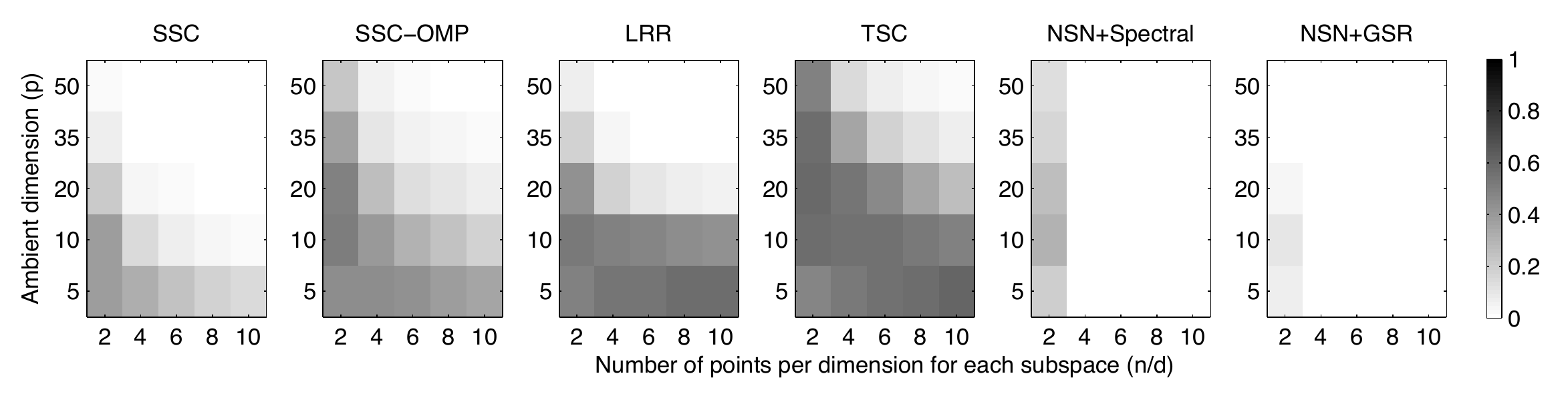}
	\caption{CE of algorithms on $5$ random $d$-dimensional subspaces and $n$ random points on each subspace. The figures shows CE for different numbers of $n/d$ and ambient dimension $p$. $d/p$ is fixed to be $3/5$. Brighter cells represent that less data points are clustered incorrectly.}
	\label{fig:synth_noiseless_random_ce}
\end{center}
\end{figure}


\begin{figure}[t]
\begin{center}
	\includegraphics[width=\textwidth]{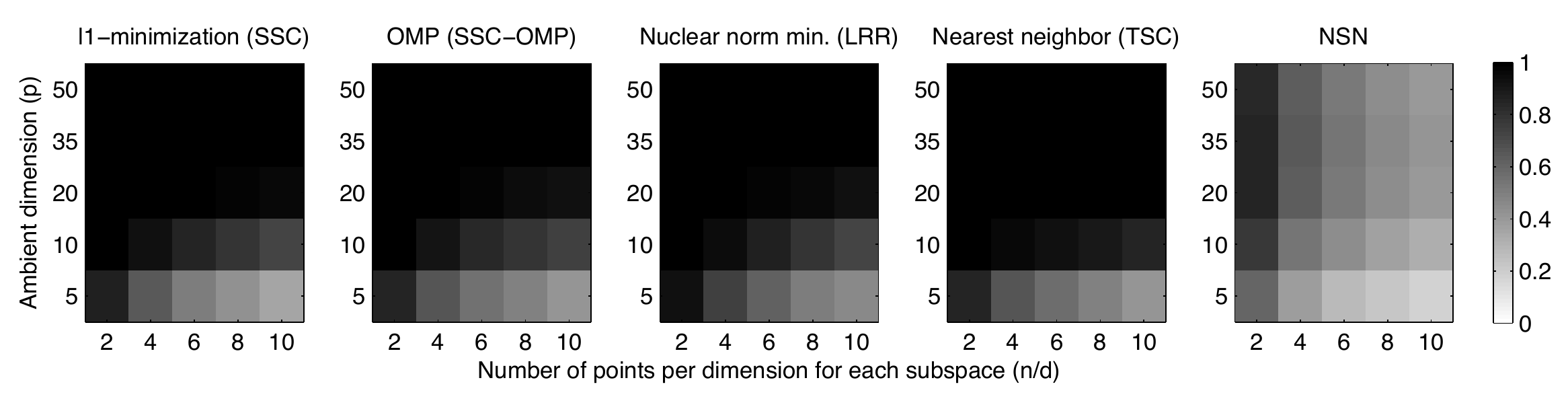}
	\caption{NSE for the same model parameters as those in Figure \ref{fig:synth_noiseless_random_ce}. Brighter cells represent that more data points have all correct neighbors.}
	\label{fig:synth_noiseless_random_nse}
\end{center}
\end{figure}


\begin{figure}[t]
\begin{center}
	\includegraphics[width=\textwidth]{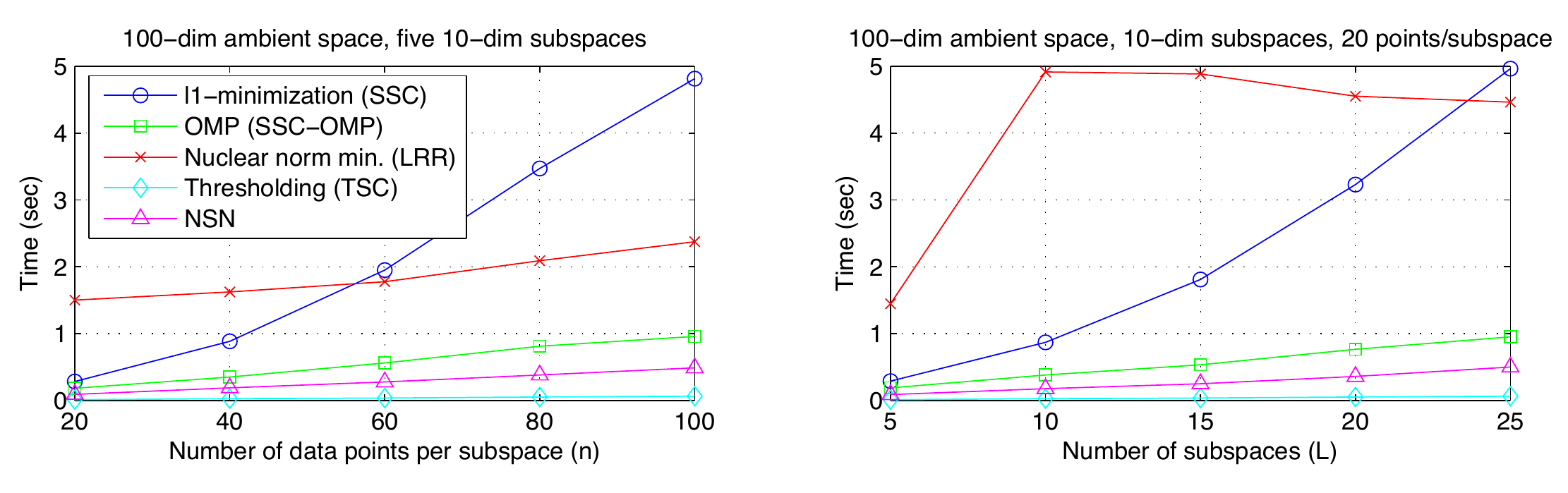}
	\caption{Average computational time of the neighborhood selection algorithms}
	\label{fig:synth_noiseless_random_et}
\end{center}
\end{figure}

We compare the performances on synthetic data generated from the fully random model. In $\mathbb{R}^p$, five $d$-dimensional subspaces are generated uniformly at random. Then for each subspace $n$ unit-norm points are generated iid uniformly at random on the subspace. To see the agreement with the theoretical result, we ran the algorithms under fixed $d/p$ and varied $n$ and $d$. We set $d/p = 3/5$ so that each pair of subspaces has intersection. Figures \ref{fig:synth_noiseless_random_ce} and \ref{fig:synth_noiseless_random_nse} show CE and NSE, respectively. Each error value is averaged over $100$ trials. Figure \ref{fig:synth_noiseless_random_ce} indicates that our algorithm clusters the data points better than the other algorithms. As predicted in the theorems, the clustering performance improves as the number of points increases. However, it also improves as the dimension of subspaces grows in contrast to the theoretical analysis. We believe that this is because our analysis on GSR is not tight. In Figure \ref{fig:synth_noiseless_random_nse}, we can see that more data points obtain correct neighbors as $n$ increases or $d$ decreases, which conforms the theoretical analysis.

We also compare the computational time of the neighborhood selection algorithms for different numbers of subspaces and data points. As shown in Figure \ref{fig:synth_noiseless_random_et}, the greedy algorithms (OMP, Thresholding, and NSN) are significantly more scalable than the convex optimization based algorithms ($\ell_1$-minimization and nuclear norm minimization).

\subsection{Real-world data : motion segmentation and face clustering}


We compare our algorithm with the existing ones in the applications of motion segmentation and face clustering. For the motion segmentation, we used Hopkins155 dataset \cite{Hopkins155}, which contains $155$ video sequences of $2$ or $3$ motions. For the face clustering, we used Extended Yale B dataset with cropped images from \citep{georghiades2001ExtYaleB,lee2005ExtYaleBcropped}. The dataset contains $64$ images for each of $38$ individuals in frontal view and different illumination conditions. To compare with the existing algorithms, we used the set of $48 \times 42$ resized raw images provided by the authors of \cite{elhamifar2013TPAMI}. The parameters of the existing algorithms were set as provided in their source codes.\footnote{As SSC-OMP and TSC do not have proposed number of parameters for motion segmentation, we found the numbers minimizing the mean CE. The numbers are given in the table.} Tables \ref{table:hopkins155_errorbysubspaces} and \ref{table:yaleB_errorbysubspaces} show CE and average computational time.\footnote{The LRR code provided by the author did not perform properly with the face clustering dataset that we used. We did not run NSN+GSR since the data points are not well distributed in its corresponding subspaces.} We can see that NSN+Spectral performs competitively with the methods with the lowest errors, but much faster. Compared to the other greedy neighborhood construction based algorithms, SSC-OMP and TSC, our algorithm performs significantly better.

\begin{table}[t]
\begin{scriptsize}
\begin{center}

	\begin{tabular}{|c||r|c|c|c|c|c|c|c|c|c|c|}

\hline
         $L$ & Algorithms & $K$-means & $K$-flats & SSC & LRR & SCC & SSC-OMP(8) & TSC(10) & NSN+Spectral(5) \\
\hline
\hline
             &     Mean CE ($\%$) & 19.80 & 13.62 & 1.52 &  2.13 &  2.06 & 16.92 & 18.44 & 3.62 \\
           2 &   Median CE ($\%$) & 17.92 & 10.65 & 0.00 &  0.00 &  0.00 & 12.77 & 16.92 & 0.00 \\
			 &    Avg. Time (sec) &   -   &  0.80 & 3.03 &  3.42 &  1.28 &  0.50 &  0.50 & 0.25 \\
\hline
             &     Mean CE ($\%$) & 26.10 & 14.07 & 4.40 &  4.03 &  6.37 & 27.96 & 28.58 & 8.28 \\
           3 &   Median CE ($\%$) & 20.48 & 14.18 & 0.56 &  1.43 &  0.21 & 30.98 & 29.67 & 2.76 \\
			 &    Avg. Time (sec) &   -   &  1.89 & 5.39 &  4.05 &  2.16 &  0.82 &  1.15 & 0.51 \\
\hline

	\end{tabular}
	\caption{CE and computational time of algorithms on Hopkins155 dataset. $L$ is the number of clusters (motions). The numbers in the parentheses represent the number of neighbors for each point collected in the corresponding algorithms.}
	\label{table:hopkins155_errorbysubspaces}

\end{center}
\end{scriptsize}
\end{table}


\begin{table}[t]
\begin{scriptsize}
\begin{center}

	\begin{tabular}{|c||r|c|c|c|c|c|c|c|}

\hline
$L$ & Algorithms & $K$-means & $K$-flats & SSC & SSC-OMP & TSC & NSN+Spectral \\
\hline
\hline
    &     Mean CE ($\%$) & 45.98 & 37.62 & 1.77 & 4.45 & 11.84 & 1.71 \\
  2 &   Median CE ($\%$) & 47.66 & 39.06 & 0.00 &  1.17 &  1.56 & 0.78 \\
	&    Avg. Time (sec) &   -   & 15.78 & 37.72 & 0.45 & 0.33 &  0.78 \\
\hline
    &     Mean CE ($\%$) & 62.55 & 45.81 & 5.77 & 6.35 & 20.02 & 3.63 \\
  3 &   Median CE ($\%$) & 63.54 & 47.92 & 1.56 & 2.86 & 15.62 & 3.12 \\
	&    Avg. Time (sec) &   -   & 27.91 & 49.45 & 0.76 & 0.60 & 3.37 \\
\hline
    &     Mean CE ($\%$) & 73.77 & 55.51 & 4.79 & 8.93 & 11.90 &  5.81 \\
  5 &   Median CE ($\%$) & 74.06 & 56.25 & 2.97 & 5.00 & 33.91 &  4.69 \\
	&    Avg. Time (sec) &   -   & 52.90 & 74.91 & 1.41 & 1.17 &  5.62 \\
\hline
    &     Mean CE ($\%$) & 79.92 & 60.12 & 7.75 & 12.90 & 38.55 &  8.46 \\
  8 &   Median CE ($\%$) & 80.18 & 60.64 & 5.86 & 12.30 & 40.14 &  7.62 \\
	&    Avg. Time (sec) &   -   & 101.3 & 119.5 & 2.84 & 2.24 & 11.51 \\
\hline
    &     Mean CE ($\%$) & 82.68 & 62.72 & 9.43 & 15.32 & 39.48 & 9.82 \\
 10 &   Median CE ($\%$) & 82.97 & 62.89 & 8.75 & 17.11 & 39.45 & 9.06 \\
	&    Avg. Time (sec) &   -   & 134.0 & 157.5 & 5.26 & 3.17 & 14.73 \\
\hline

	\end{tabular}
	\caption{CE and computational time of algorithms on Extended Yale B dataset. For each number of clusters (faces) $L$, the algorithms ran over 100 random subsets drawn from the overall 38 clusters.}
	\label{table:yaleB_errorbysubspaces}	

\end{center}
\end{scriptsize}
\end{table}

\newpage
\bibliography{greedysc}
\bibliographystyle{plainnat}

\newpage
\appendix

\section{Discussion on implementation issues}

\subsection{A faster implementation for NSN} \label{sec:fastimpl}
At each step of NSN, the algorithm computes the projections of all points onto a subspace and find one with the largest norm. A naive implementation of the algorithm requires $O(pK^2N^2)$ time complexity.

In fact, we can reduce the complexity to $O(pKN^2)$. Instead of finding the maximum norm of the projections, we can find the maximum squared norm of the projections. Let $\set{U}_k$ be the subspace $\set{U}$ at step $k$. For any data point $y$, we have
$$
\|\proj_{\set{U}_k} y\|_2^2 = \|\proj_{\set{U}_{k-1}} y\|_2^2 + |u_k^\top y|^2
$$
where $u_k$ is the new orthogonal axis added from $\set{U}_{k-1}$ to make $\set{U}_k$. That is, $\set{U}_{k-1} \perp u_k$ and $\set{U}_k = \set{U}_{k-1} \bigoplus u_k$. As $\|\proj_{\set{U}_{k-1}} y\|_2^2$ is already computed in the $(k-1)$'th step, we do not need to compute it again at step $k$. Based on this fact, we have a faster implementation as described in the following. Note that $P_j$ at the $k$th step is equal to $\|\proj_{\set{U}_k} y_j\|_2^2$ in the original NSN algorithm.
\begin{algorithm}
	\caption{Fast Nearest Subspace Neighbor (F-NSN)}
\begin{algorithmic} \label{alg:fnsn}
	\Require A set of $N$ samples $\mathcal{Y} = \{y_1,\ldots,y_N\}$, The number of required neighbors $K$, Maximum subspace dimension $k_{max}$.
	\Ensure A neighborhood matrix $W \in \{0,1\}^{N \times N}$
	\State $y_i \gets y_i / \|y_i\|_2$, $\forall i \in [N]$
	\For {$i = 1,\ldots,N$}
		\State $\set{I}_i \gets \{i\}$, $u_1 \gets y_i$
		\State $P_j \gets 0, \forall j \in [N]$
		\For {$k = 1,\ldots,K$}
			\If {$k \le k_{max}$}
				\State $P_j \gets P_j + {\|u_k^\top y_j\|^2} ,~ \forall j \in [N]$
			\EndIf
			\State $j^* \gets \arg\max_{j \in [N], j \notin \set{I}_i} P_j$
			\State $\set{I}_i \gets \set{I}_i \cup \{j^*\}$
			\If {$k < k_{max}$}
				\State $u_{k+1} \gets \frac{y_{j^*} - \sum_{l=1}^k (u_l^\top y_{j^*}) u_l}{\|y_{j^*} - \sum_{l=1}^k (u_l^\top y_{j^*}) u_l\|_2}$
			\EndIf
		\EndFor
		\State $W_{ij} \gets \mathbb{I}_{j \in \set{I}_i \text{ or } P_j = 1} ,~ \forall j \in [N]$
	\EndFor
\end{algorithmic}
\end{algorithm}

\subsection{Estimation of the number of clusters}
When $L$ is unknown, it can be estimated at the clustering step. For Spectral clustering, a well-known approach to estimate $L$ is to find a “knee” point in the singular values of the neighborhood matrix. It is the point where the difference between two consecutive singular values are the largest. For GSR, we do not need to estimate the number of clusters a priori. Once the algorithms finishes, the number of the resulting groups will be our estimate of $L$.

\subsection{Parameter setting}
The choices of $K$ and $k_{\max}$ depend on the dimension of the subspaces $d$. If data points are lying exactly on the model subspaces, $K=k_{\max}=d$ is enough for GSR to recover a subspace. In practical situations where the points are near the subspaces, it is better to set $K$ to be larger than $d$. $k_{\max}$ can also be larger than $d$ because the $k_{\max}-d$ additional dimensions, which may be induced from the noise, do no intersect with the other subspaces in practice. For Extended Yale B dataset and Hopkins155 dataset, we found that NSN+Spectral performs well if $K$ is set to be around $2d$.

\section{Proofs}

\subsection{Proof outline} \label{sec:proofoutline}

We describe the first few high-level steps in the proofs of our main theorems. Exact clustering of our algorithms depends on whether NSN can find all correct neighbors for the data points so that the following algorithm (GSR or Spectral clustering) can cluster the points exactly. For NSN+GSR, exact clustering is guaranteed when there is a point on each subspace that have all correct neighbors which are at least $d-1$. For NSN+Spectral, exact clustering is guaranteed when each data point has only the $n-1$ other points on the same subspace as neighbors. In the following, we explain why these are true.

\vspace{0.1in}
\noindent \underline{\textbf{Step 1-1: Exact clustering condition for GSR}}
\vspace{0.1in}

The two statistical models have a property that for any $d$-dimensional subspace in $\mathbb{R}^p$ other than the true subspaces $\set{D}_1,\ldots,\set{D}_L$ the probability of any points lying on the subspace is zero. Hence, we claim the following.
\begin{fact}[Best $d$-dimensional fit]
With probability one, the true subspaces $\set{D}_1,\ldots,\set{D}_L$ are the $L$ subspaces containing the most points among the set of all possible $d$-dimensional subspaces.
\end{fact}
Then it suffices for each subspace to have one point whose neighbors are $d-1$ all correct points on the same subspace. This is because the subspace spanned by those $d$ points is almost surely identical to the true subspace they are lying on, and that subspace will be picked by GSR.
\begin{fact} \label{thm:atleastonepoint}
If NSN with $K \ge d-1$ finds all correct neighbors for at least one point on each subspace, GSR recovers all the true subspaces and clusters the data points exactly with probability one. 
\end{fact}
In the following steps, we consider one data point for each subspace. We show that NSN with $K = k_{max} = d$ finds all correct neighbors for the point with probability at least $1-\frac{3\delta}{1-\delta}$. Then the union bound and Fact \ref{thm:atleastonepoint} establish exact clustering with probability at least $1 - \frac{3L\delta}{1-\delta}$.

\vspace{0.1in}
\noindent \underline{\textbf{Step 1-2: Exact clustering condition for spectral clustering}}
\vspace{0.1in}

It is difficult to analyze spectral clustering for the resulting neighborhood matrix of NSN. A trivial case for a neighborhood matrix to result in exact clustering is when the points on the same subspace form a single fully connected component. If NSN with $K = k_{max} = d$ finds all correct neighbors for every data point, the subspace $\set{U}$ at the last step ($k = K$) is almost surely identical to the true subspace that the points lie on. Hence, the resulting neighborhood matrix $W$ form $L$ fully connected components each of which contains all of the points on the same subspace.

In the rest of the proof, we show that if \eqref{eqn:NSNfullyrandom} holds, NSN finds all correct neighbors for a fixed point with probability $1-\frac{3\delta}{1-\delta}$. Let us assume that this is true. If \eqref{eqn:NSNfullyrandom} with $C_1$ and $C_2$ replaced by $\frac{C_1}{4}$ and $\frac{C_2}{2}$ holds, we have
\begin{align*}
n > C_1 d \left( \log \frac{ne}{d(\delta/n)} \right)^2 ,\quad \frac{d}{p} < \frac{C_2 \log n}{\log (ndL (\delta/n)^{-1})}.
\end{align*}
Then it follows from the union bound that NSN finds all correct neighbors for all of the $n$ points on each subspace with probability at least $1 - \frac{3L\delta}{1-\delta}$, and hence we obtain $W_{ij} = \mathbb{I}_{w_i = w_j}$ for every $(i,j) \in [N]^2$. Exact clustering is guaranteed.

\vspace{0.1in}
\noindent \underline{\textbf{Step 2: Success condition for NSN}}
\vspace{0.1in}

Now the only proposition that we need to prove is that for each subspace $\set{D}_i$ NSN finds all correct neighbors for a data point (which is a uniformly random unit vector on the subspace) with probability at least $1-\frac{3\delta}{1-\delta}$. As our analysis is independent of the subspaces, we only consider $\set{D}_1$. Without loss of generality, we assume that $y_1$ lies on $\set{D}_1$ ($w_1 = 1$) and focus on this data point.

When NSN finds neighbors of $y_1$, the algorithm constructs $k_{max}$ subspaces incrementally. At each step $k = 1,\ldots, K$, if the largest projection onto $\set{U}$ of the uncollected points on the same true subspace $\set{D}_1$ is greater than the largest projection among the points on different subspaces, then NSN collects a correct neighbor. In a mathematical expression, we want to satisfy
\begin{align} \label{eqn:NSNgeneralsuccesscond0}
\max_{j : w_j = 1, j \notin \set{I}_1} \|\proj_{\set{U}} y_j\|_2 &> \max_{j : w_j \neq 1, j \notin \set{I}_1} \|\proj_{\set{U}} y_j\|_2
\end{align}
for each step of $k = 1, \ldots, K$.

The rest of the proof is to show \eqref{eqn:NSNfullyrandom} and \eqref{eqn:NSNsemirandom} lead to \eqref{eqn:NSNgeneralsuccesscond0} with probability $1-\frac{3\delta}{1-\delta}$ in their corresponding models. It is difficult to prove \eqref{eqn:NSNgeneralsuccesscond0} itself because the subspaces, the data points, and the index set $\set{I}_1$ are all dependent of each other. Instead, we introduce an Oracle algorithm whose success is equivalent to the success of NSN, but the analysis is easier. Then the Oracle algorithm is analyzed using stochastic ordering, bounds on order statistics of random projections, and the measure concentration inequalities for random subspaces. The rest of the proof is provided in Sections \ref{sec:prooftechnical_semirandom} and \ref{sec:prooftechnical_fullyrandom}.

\subsection{Preliminary lemmas}

Before we step into the technical parts of the proof, we introduce the main ingredients which will be used. The following lemma is about upper and lower bounds on the order statistics for the projections of iid uniformly random unit vectors.

\begin{lemma} \label{lem:projbound_unitvec}
Let $x_1, \ldots, x_n$ be drawn iid uniformly at random from the $d$-dimensional unit ball $\mathbb{S}^{d-1}$. Let $z_{(n-m+1)}$ denote the $m$'th largest value of $\{z_i \triangleq \|Ax_i\|_2, 1 \le i \le n\}$ where $A \in \mathbb{R}^{k \times d}$.
\begin{enumerate}
\item[a.] Suppose that the rows of $A$ are orthonormal to each other. For any $\alpha \in (0,1)$, there exists a constant $C > 0$ such that for $n,m,d,k \in \mathbb{N}$ where
\begin{align}
n-m+1 \ge C m \left( \log \frac{ne}{m\delta} \right)^2 \label{eqn:projbound_condition}
\end{align}
we have
\begin{align}
z_{(n-m+1)}^2 > \frac{k}{d} + \frac{1}{d} \cdot \min \left\{ 2 \log \left( \frac{n-m+1}{C m \left( \log \frac{ne}{m\delta} \right)^2 } \right) , \alpha \sqrt{d-k} \right\} \label{eqn:projbound_unitvec1}
\end{align}
with probability at least $1 - \delta^m$.
\item[b.] For any $k \times d$ matrix $A$,
\begin{align}
z_{(n-m+1)} < \frac{\|A\|_F}{\sqrt{d}} + \frac{\|A\|_2}{\sqrt{d}} \cdot \left( \sqrt{2\pi} + \sqrt{ 2 \log \frac{ne}{m \delta}} \right)  \label{eqn:projbound_unitvec3}
\end{align}
with probability at least $1 - \delta^m$.
\end{enumerate}
\end{lemma}


Lemma \ref{lem:projbound_unitvec}b can be proved by using the measure concentration inequalities \cite{ledoux2005concentration}. Not only can they provide inequalities for random unit vectors, they also give us inequalities for random subspaces.

\begin{lemma} \label{lem:subspace_concentration}
Let the columns of $X \in \mathbb{R}^{d \times k}$ be the orthonormal basis of a $k$-dimensional random subspace drawn uniformly at random in $d$-dimensional space.
\begin{enumerate}
\item[a.] For any matrix $A \in \mathbb{R}^{p \times d}$.
$$
E[\|AX\|_F^2] = \frac{k}{d} \|A\|_F^2
$$

\item[b.] \citep{milman1986asymptotic, ledoux2005concentration}
If $\|A\|_2$ is bounded, then we have
\begin{align*}
\Pr \left\{ \|A X\|_F > \sqrt{\frac{k}{d}} \|A\|_F + \|A\|_2 \cdot \left( \sqrt{\frac{8\pi}{d-1}} + t \right) \right\} \le e^{-\frac{(d-1)t^2}{8}}.
\end{align*}
\end{enumerate}
\end{lemma}

\subsection{Proof of Theorem \ref{thm:NSNsemirandom}} \label{sec:prooftechnical_semirandom}

Following Section \ref{sec:proofoutline}, we show in this section that if \eqref{eqn:NSNsemirandom} holds then NSN finds all correct neighbors for $y_1$ (which is assumed to be on $\set{D}_1$) with probability at least $1-\frac{3\delta}{1-\delta}$. 

\vspace{0.1in}
\noindent \underline{\textbf{Step 3: NSN Oracle algorithm}}
\vspace{0.1in}

Consider the Oracle algorithm in the following. Unlike NSN, this algorithm knows the true label of each data point. It picks the point closest to the current subspace among the points with the same label. Since we assume $w_1 = 1$, the Oracle algorithm for $y_1$ picks a point in $\{y_j : w_j = 1\}$ at every step.
\begin{algorithm}
\caption{NSN Oracle algorithm for $y_1$ (assuming $w_1 = 1$)}
\begin{algorithmic} \label{algo:nsnsc_oracle}
	\Require A set of $N$ samples $\mathcal{Y} = \{y_1,\ldots,y_N\}$, The number of required neighbors $K = d-1$, Maximum subspace dimension $k_{max} = \lceil 2 \log d \rceil$
	\State $\set{I}_1^{(1)} \gets \{1\}$
	\For {$k = 1,\ldots,K$}
		\If {$k \le k_{max}$}
			\State $\set{V}_k \gets \sp\{y_j:j \in \set{I}_1^{(k)}\}$
			\State $j_k^* \gets \arg\max_{j \in [N]: w_j = 1, j \notin \set{I}_1^{(k)}} \|\proj_{\set{V}_k} y_j\|_2$
		\Else
			\State $j_k^* \gets \arg\max_{j \in [N]: w_j = 1, j \notin \set{I}_1^{(k)}} \|\proj_{\set{V}_{k_{max}}} y_j\|_2$
		\EndIf

		\If {$\max_{j \in [N]: w_j = 1, j \notin \set{I}_i^{(k)}} \|\proj_{\set{V}_k} y_j\|_2 \le \max_{j \in [N]: w_j \neq 1} \|\proj_{\set{V}_k} y\|_2$}
			\State Return FAILURE
		\EndIf

		\State $\set{I}_1^{(k+1)} \gets \set{I}_1^{(k)} \cup \{j_k^*\}$
	\EndFor
	\State Return SUCCESS
\end{algorithmic}
\end{algorithm}

Note that the Oracle algorithm returns failure if and only if the original algorithm picks an incorrect neighbor for $y_1$. The reason is as follows. Suppose that NSN for $y_1$ picks the first incorrect point at step $k$. By the step $k-1$, correct points have been chosen because they are the nearest points for the subspaces in the corresponding steps. The Oracle algorithm will also pick those points because they are the nearest points among the correct points. Hence $\set{U} \equiv \set{V}_k$. At step $k$, NSN picks an incorrect point as it is the closest to $\set{U}$. The Oracle algorithm will declare failure because that incorrect point is closer than the closest point among the correct points. In the same manner, we see that NSN fails if the Oracle NSN fails. Therefore, we can instead analyze the success of the Oracle algorithm. The success condition is written as
\begin{align} 
\|\proj_{\set{V}_k} y_{j_k^*}\|_2 &> \max_{j \in [N]: w_j \neq 1} \|\proj_{\set{V}_k} y\|_2 ,\quad \forall k = 1,\ldots,k_{max}, \nonumber \\
\|\proj_{\set{V}_{k_{max}}} y_{j_k^*}\|_2 &> \max_{j \in [N]: w_j \neq 1} \|\proj_{\set{V}_{k_{max}}} y\|_2 ,\quad \forall k = k_{max}+1,\ldots,K. \label{eqn:NSNgeneralsuccesscond2}
\end{align}
Note that $\set{V}_k$'s are independent of the points $\{y_j : j \in [N], w_j \neq 1\}$. We will use this fact in the following steps.

\vspace{0.1in}
\noindent \underline{\textbf{Step 4: Lower bounds on the projection of correct points (the LHS of \eqref{eqn:NSNgeneralsuccesscond2})}}
\vspace{0.1in}

Let $V_k \in \mathbb{R}^{d \times k}$ be such that the columns of $D_1 V_k$ form an orthogonal basis of $\set{V}_k$. Such a $V_k$ exists because $\set{V}_k$ is a $k$-dimensional subspace of $\set{D}_1$. Then we have
$$
\|\proj_{\set{V}_k} y_{j_k^*}\|_2 = \|V_k^\top D_1^\top D_1 x_{j_k^*}\|_2 = \|V_k^\top x_{j_k^*}\|_2
$$
In this step, we obtain lower bounds on $\|V_k^\top x_{j_k^*}\|_2$ for $k \le k_{max}$ and $\|V_{k_{max}}^\top x_{j_k^*}\|_2$ for $k > k_{max}$.

It is difficult to analyze $\|V_k^\top x_{j_k^*}\|_2$ because $V_k$ and $x_{j_k^*}$ are dependent. We instead analyze another random variable that is stochastically dominated by $\|V_k^\top x_{j_k^*}\|_2^2$. Then we use a high-probability lower bound on that variable which also lower bounds $\|V_k^\top x_{j_k^*}\|_2^2$ with high probability.

Define $P_{k,(m)}$ as the $m$'th largest norm of the projections of $n-1$ iid uniformly random unit vector on $\mathbb{S}^{d-1}$ onto a $k$-dimensional subspace. Since the distribution of the random unit vector is isotropic, the distribution of $P_{k,(m)}$ is identical for any $k$-dimensional subspaces independent of the random unit vectors. We have the following key lemma.
\begin{lemma} \label{lem:NSNstochasticdominance}
$\|V_k^\top x_{j_k^*}\|_2$ stochastically dominates $P_{k,(k)}$, i.e.,
$$
\Pr \{\|V_k^\top x_{j_k^*}\|_2 \ge t \} \ge \Pr \{ P_{k,(k)} \ge t \}
$$
for any $t \ge 0$. Moreover, $P_{k,(k)} \ge P_{\hat{k},(k)}$ for any $\hat{k} \le k$. 
\end{lemma}
The proof of the lemma is provided in Appendix \ref{sec:pf_stocordering}. Now we can use the lower bound on $P_{k,(k)}$ given in Lemma \ref{lem:projbound_unitvec}a to bound $\|V_k^\top x_{j_k^*}\|_2$. Let us pick $\alpha$ and $C$ for which the lemma holds. The first inequality of \eqref{eqn:NSNsemirandom} with $C_1 = C + 1$ leads to $n-d > C d \left( \log \frac{ne}{d\delta} \right)^2$, and also
\begin{align}
n-k > C k \left( \log \frac{ne}{k\delta} \right)^2 ,\quad \forall k = 1,\ldots,d-1. \label{eqn:lemma_condition}
\end{align}
Hence, it follow from Lemma \ref{lem:projbound_unitvec}a that for each $k = 1, \ldots, k_{max}$, we have
\begin{align}
\|V_k^\top x_{j_k^*}\|_2 
&\ge \frac{k}{d} + \frac{1}{d} \min\left\{ 2\log\left(\frac{n-k+1}{Ck\left(\log\frac{ne}{k\delta}\right)^2}\right), \alpha\sqrt{d-k} \right\} \nonumber \\
&\ge \frac{k}{d} + \frac{1}{d} \min\left\{ 2\log\left(\frac{n-d}{Cd\left(\log\frac{ne}{\delta}\right)^2}\right), \alpha\sqrt{d - 2 \log d} \right\} \label{eqn:lowerbound1}
\end{align}
with probability at least $1-\delta^k$.

For $k > k_{max}$, we want to bound $\|\proj_{\set{V}_{k_{max}}} y_{j_k^*}\|_2$. We again use Lemma \ref{lem:NSNstochasticdominance} to obtain the bound. Since the condition for the lemma holds as shown in \eqref{eqn:lemma_condition}, we have
\begin{align} 
\|V_{k_{max}}^\top x_{j_k^*}\|_2
&\ge \frac{2 \log d}{d} + \frac{1}{d} \min\left\{ 2\log\left(\frac{n-k+1}{Ck\left(\log\frac{ne}{k\delta}\right)^2}\right), \alpha\sqrt{d - 2 \log d} \right\} \nonumber \\
&\ge \frac{2 \log d}{d} + \frac{1}{d} \min\left\{ 2\log\left(\frac{n-d}{Cd\left(\log\frac{ne}{\delta}\right)^2}\right), \alpha\sqrt{d - 2 \log d} \right\} \label{eqn:lowerbound2}
\end{align}
with probability at least $1-\delta^k$, for every $k = k_{max}+1,\ldots,d-1$.

The union bound gives that \eqref{eqn:lowerbound1} and \eqref{eqn:lowerbound2} hold for all $k = 1,\ldots,d-1$ simultaneously with probability at least $1-\frac{\delta}{1-\delta}$.
\\

\vspace{0.1in}
\noindent \underline{\textbf{Step 5: Upper bounds on the projection of incorrect points (the RHS of \eqref{eqn:NSNgeneralsuccesscond2})}}
\vspace{0.1in}

Since we have $\|\proj_{\set{V}_k} y_j\|_2 = \|V_k^\top D_1^\top D_{w_j} x_j\|_2$, the RHS of \eqref{eqn:NSNgeneralsuccesscond2} can be written as
\begin{align} \label{eqn:NSNconditionRHS}
\max_{j : j \in [N], w_j \neq 1} \|V_k^\top D_1^\top D_{w_j} x_j\|_2
\end{align}
In this step, we want to bound \eqref{eqn:NSNconditionRHS} for every $k=1,\ldots,d-1$ by using the concentration inequalities for $V_k$ and $x_j$. Since $V_k$ and $x_j$ are independent, the inequality for $x_j$ holds for any fixed $V_k$.

It follows from Lemma \ref{lem:projbound_unitvec}b and the union bound that with probability at least $1-\delta$,
\begin{align*}
&\max_{j : j \in [N], w_j \neq 1} \|V_k^\top D_1^\top D_{w_j} x_j\|_2 \\
&\le \frac{ \max_{l \neq 1} \|V_k^\top D_1^\top D_l \|_F}{\sqrt{d}} + \frac{\max_{l \neq 1} \|V_k^\top D_1^\top D_l \|_2}{\sqrt{d}} \cdot \left( \sqrt{2\pi} + \sqrt{ 2 \log \frac{n(L-1)e}{\delta/d}} \right) \\
&\le \frac{ \max_{l \neq 1} \|V_k^\top D_1^\top D_l \|_F}{\sqrt{d}} \cdot \left( 5 + \sqrt{ 2 \log \frac{ndL}{\delta}} \right) 
\end{align*}
for all $k = 1,\ldots,d-1$. The last inequality follows from the fact $\|V_k^\top D_1^\top D_l\|_2 \le \|V_k^\top D_1^\top D_l\|_F$. Since $\|V_k^\top D_1^\top D_{w_j} x_j\|_2 \le \|V_k^\top D_1^\top D_{w_j}\|_F \le \max_{l \neq 1} \|V_k^\top D_1^\top D_l\|_F$ for every $j$ such that $w_j \neq 1$, we have
\begin{align}
\max_{j : j \in [N], w_j \neq 1} \|V_k^\top D_1^\top D_{w_j} x_j\|_2 &\le \frac{ \max_{l \neq 1} \|V_k^\top D_1^\top D_l \|_F}{\sqrt{d}} \cdot \min \left\{ 5 + \sqrt{ 2 \log \frac{ndL}{\delta}}, \sqrt{d} \right\}. 
\label{eqn:NSNconcentration2}
\end{align}

Now let us consider $\max_{l \neq 1} \|V_k^\top D_1^\top D_l \|_F$. In our statistical model, the new axis added to $\set{V}_k$ at the $k$th step ($u_{k+1}$ in Algorithm 3) is chosen uniformly at random from the subspace in $\set{D}_1$ orthogonal to $\set{V}_k$. Therefore, $V_k$ is a random matrix drawn uniformly from the $d \times k$ Stiefel manifold, and the probability measure is the normalized Haar (rotation-invariant) measure. From Lemma \ref{lem:subspace_concentration}b and the union bound, we obtain that with probability at least $1-\delta/dL$,
\begin{align}
\|V_k^\top D_1^\top D_l\|_F
&\le \sqrt{\frac{k}{d}} \|D_1^\top D_l\|_F + \|D_1^\top D_l\|_2 \cdot \left( \sqrt{\frac{8\pi}{d-1}} + \sqrt{ \frac{8}{d-1} \log \frac{ dL}{\delta}}\right) \nonumber \\
&\le \|D_1^\top D_l\|_F \cdot \left( \sqrt{\frac{k}{d}} + \sqrt{\frac{8\pi}{d-1}} +  \sqrt{ \frac{8}{d-1} \log \frac{dL}{\delta}} \right) \nonumber \\
&\le \max\aff \cdot \sqrt{d} \cdot \left( \sqrt{\frac{k}{d}} + \sqrt{\frac{8\pi}{d-1}} +  \sqrt{ \frac{8}{d-1} \log \frac{dL}{\delta}} \right).
\label{eqn:NSNconcentration1}
\end{align}
The union bound gives that with probability at least $1-\delta$, $\max_{l \neq 1} \|V_k^\top D_1^\top D_l \|_F$ is also bounded by \eqref{eqn:NSNconcentration1} for every $k = 1,\ldots,k_{max}$.

Putting \eqref{eqn:NSNconcentration1} and \eqref{eqn:NSNconcentration2} together, we obtain
\begin{align}
&\max_{j : j \in [N], w_j \neq 1} \|V_k^\top D_1^\top D_{w_j} x_j\|_2 \nonumber \\
&\le \max\aff \cdot \left( \sqrt{\frac{k}{d}} + \sqrt{\frac{8\pi}{d-1}} +  \sqrt{ \frac{8}{d-1} \log \frac{dL}{\delta}} \right) \cdot \min \left\{ 5 + \sqrt{ 2 \log \frac{ndL}{\delta}} , \sqrt{d} \right\} \label{eqn:NSNboundRHS} 
\end{align}
for all $k=1,\ldots,d-1$ with probability at least $1 - 2\delta$.

\vspace{0.1in}
\noindent \underline{\textbf{Final Step: Proof of the main theorem}}
\vspace{0.1in}

Putting \eqref{eqn:lowerbound1}, \eqref{eqn:lowerbound2}, and \eqref{eqn:NSNboundRHS} together, we obtain that if
\begin{align} \label{eqn:bound1}
	\max \aff < \min_{1 \le k \le d-1} \frac{ \sqrt{ \min\{k, 2 \log d\} + \min\left\{ 2\log\left(\frac{n-d}{Cd}\right) - 4 \log \log\frac{ne}{\delta}, \alpha\sqrt{d - 2 \log d} \right\}} }           
										    { \left(\sqrt{\min\{k, 2 \log d\}} + \sqrt{\frac{8\pi d}{d-1}} +  \sqrt{ \frac{8d}{d-1} \log \frac{dL}{\delta}} \right) \cdot \min \left\{ 5 + \sqrt{ 2 \log \frac{ndL}{\delta}} , \sqrt{d} \right\} },
\end{align}
then \eqref{eqn:NSNgeneralsuccesscond2} holds, and hence NSN finds all correct neighbors for $y_1$ with probability at least $1-\frac{3\delta}{1-\delta}$. The RHS of \eqref{eqn:bound1} is minimized when $k \ge 2 \log d$, and consequently the condition \eqref{eqn:bound1} is equivalent to
\begin{align} \label{eqn:bound2}
	\max \aff < \frac{ \sqrt{ 2 \log d + \min\left\{ 2\log\left(\frac{n-d}{Cd}\right) - 4 \log \log\frac{ne}{\delta}, \alpha\sqrt{d - 2 \log d} \right\}} }           
										    { \left(\sqrt{2 \log d} + \sqrt{\frac{8\pi d}{d-1}} +  \sqrt{ \frac{8d}{d-1} \log \frac{dL}{\delta}} \right) \cdot \min \left\{ 5 + \sqrt{ 2 \log \frac{ndL}{\delta}} , \sqrt{d} \right\} }.
\end{align}
As $n$ is polynomial in $d$, there is a constant $C_3>0$ such that 
\begin{align*}
	\text{(RHS of \eqref{eqn:bound2})} > \frac{C_3 \sqrt{\log\left(n-d\right) - \log\log \frac{ne}{\delta} }}{\sqrt{\log \frac{dL}{\delta} \cdot \log \frac{ndL}{\delta}}}
\end{align*}
This completes the proof.

\subsection{Proof of Theorem \ref{thm:NSNfullyrandom}} \label{sec:prooftechnical_fullyrandom}
As we did in Section \ref{sec:prooftechnical_semirandom}, we prove in this section that if \eqref{eqn:NSNfullyrandom} holds then NSN finds all correct neighbors for $y_1$ with probability at least $1-\frac{3\delta}{1-\delta}$. 

The only difference between the semi-random model and the fully random model is the statistical dependence between subspaces. We can follow Step 3 in Section \ref{sec:prooftechnical_semirandom} because they do not take any statistical dependence between subspaces into account. We assert that \eqref{eqn:NSNgeneralsuccesscond2} is the success condition also for the fully random model. However, as $K = k_{max} = d$, there is no case where $k > k_{max}$ in this proof.

Now we provide a new proof of the last three steps for the fully random model.

\vspace{0.1in}
\noindent \underline{\textbf{Step 4: Lower bounds on the projection of correct points (the LHS of \eqref{eqn:NSNgeneralsuccesscond2})}}
\vspace{0.1in}

We again use Lemma \ref{lem:NSNstochasticdominance}. For $k > d/2$, we use the fact that $\|V_k^\top x_{j_k^*}\|_2$ stochastically dominates $P_{\lfloor d/2 \rfloor, (k)}$. Then it follows from Lemma \ref{lem:projbound_unitvec}a that
\begin{align}
\|V_k^\top x_{j_k^*}\|_2 
&\ge \frac{k}{2d} + \frac{1}{d} \min\left\{ 2\log\left(\frac{n-k+1}{Ck\left(\log\frac{ne}{k\delta}\right)^2}\right), \alpha\sqrt{d/2} \right\} \label{eqn:lowerbound3}
\end{align}
for all $k = 1,\ldots,d-1$ simultaneously with probability at least $1-\frac{\delta}{1-\delta}$.

\vspace{0.1in}
\noindent \underline{\textbf{Step 5: Upper bounds on the projection of incorrect points (the RHS of \eqref{eqn:NSNgeneralsuccesscond2})}}
\vspace{0.1in}

We again use the notion of $X_k \in \mathbb{R}^{d \times k}$ which is defined in the proof of Theorem \ref{thm:NSNsemirandom}. Since $\|\proj_{\set{V}_k} y_j\|_2 = \|V_k^\top D_1^\top y_j \|_2$, the RHS of \eqref{eqn:NSNgeneralsuccesscond2} can be written as
\begin{align} \label{eqn:NSN2conditionRHS}
\max_{j : j \in [N], w_j \neq 1} \|V_k^\top D_1^\top y_j\|_2
\end{align}
Since the true subspaces are independent of each other, $y_j$ with $w_j \neq 1$ is also independent of $D_1$ and $V_k$, and its marginal distribution is uniform over $\mathbb{S}^{p-1}$. It follows from Lemma \ref{lem:projbound_unitvec}b that with probability at least $1-\delta/d$,
\begin{align}
\eqref{eqn:NSN2conditionRHS} &\le \frac{ \|V_k^\top D_1^\top\|_F}{\sqrt{p}} + \frac{\|V_k^\top D_1^\top\|_2}{\sqrt{p}} \cdot \sqrt{ 2 \log \frac{n(L-1)e}{\delta/d}} \nonumber \\
&\le \sqrt{\frac{k}{p}} + \sqrt{ \frac{2}{p} \log \frac{ndLe}{\delta}}.
\label{eqn:NSN2concentration1}
\end{align}
The last inequality is obtained using the facts $\|D_1 V_k\|_F = \sqrt{k}$ and $\|D_1 V_k\|_2 \le 1$. The union bound provides that \eqref{eqn:NSN2concentration1} holds for every $k=1,\ldots,d-1$ with probability at least $1-\delta$.

\vspace{0.1in}
\noindent \underline{\textbf{Final Step: Proof of the main theorem}}
\vspace{0.1in}

Now it suffices to show that $\eqref{eqn:lowerbound3}>\eqref{eqn:NSN2concentration1}$ for every $k = 1,2,\ldots,d-1$, i.e.,
\begin{align}
 \sqrt{\frac{k}{2d} + \frac{1}{d} \min\left\{ 2\log\left(\frac{n-k+1}{Ck\left(\log\frac{ne}{k\delta}\right)^2}\right), \alpha\sqrt{\frac{d}{2}} \right\}}  > \sqrt{\frac{k}{p}} + \sqrt{ \frac{2}{p} \log \frac{ndLe}{\delta}} ,\quad \forall k=1,2,\ldots,d-1. \label{eqn:NSN2successcond}
\end{align}
where $\alpha, C$ are the constants described in Lemma \ref{lem:projbound_unitvec}a. \eqref{eqn:NSN2successcond} is equivalent to
\begin{align}
\frac{d}{p} < \min_{1 \le k \le d-1} \frac{ k/2 + \min\left\{ 2\log\left(\frac{n-k+1}{Ck}\right) - 4 \log \left( \log\frac{ne}{k\delta} \right), \alpha\sqrt{d/2} \right\} }{ \left( \sqrt{k} + \sqrt{2 \log (ndL\delta^{-1}e)} \right)^2 }. \label{eqn:NSN2successcond2}
\end{align}
As $n$ is polynomial in $d$, the numerator can be replaced by $O(k+\log(n-k+1))$. The RHS is minimized when $k = O(\log (ndL\delta^{-1}))$. Hence, the above condition is satisfied if \eqref{eqn:NSNfullyrandom} holds.

\subsection{Proof of Lemma \ref{lem:NSNstochasticdominance}} \label{sec:pf_stocordering}
We construct a generative model for two random variables that are equal in distribution to $\|V_k^\top x_{j_k^*}\|_2^2$ and $P_{k,(k)}^2$. Then we show that the one corresponding to $\|V_k^\top x_{j_k^*}\|_2^2$ is greater than the other corresponding to $P_{k,(k)}^2$. This generative model uses the fact that for any isotropic distributions the marginal distributions of the components along any orthogonal axes are invariant.
 
The generative model is given as follows.
\begin{enumerate}
\item For $k=1,\ldots,k_{max}$, repeat 2.
\item Draw $n-1$ iid random variables $Y^{(k)}_1,\ldots,Y^{(k)}_{n-1}$ as follows.
$$
Y^{(k)}_j = \left(1-\sum_{i=1}^{k-1}Y^{(i)}_j\right) \cdot (X^{(k)}_{j1})^2 ,~ X^{(k)}_j \sim \mathrm{Unif}(\mathbb{S}^{d-k}),\quad \forall j=1,\ldots,n-1.
$$
where $X_{j1}^{(k)}$ is the first coordinate of $X_j^{(k)}$.
Define
$$
\pi_k \triangleq \arg \max_{j: j \neq \pi_1,\ldots,\pi_{k-1}} \left( \sum_{i=1}^{k} Y^{(i)}_j \right).
$$

\item For $k=k_{max}+1,\ldots,d-1$, repeat
$$
\pi_k \triangleq \arg \max_{j: j \neq \pi_1,\ldots,\pi_{k-1}} \left( \sum_{i=1}^{k_{max}} Y^{(i)}_j \right).
$$

\end{enumerate}

We first claim that $(\sum_{i=1}^k Y_{\pi_k}^{(i)})$ is equal in distribution to $\|V_k^\top x_{j_k^*}\|_2^2$. Consider the following two sets of random variables.
\begin{align*}
A_k &\triangleq \left( \sum_{i=1}^{k} Y^{(i)}_j : j \in [n-1], j \neq \pi_1, \ldots, \pi_{k-1} \right) ,\\
B_k &\triangleq \left( \|V_k^\top x_j\|_2^2 : w_j = 1, j \neq 1,j_1^*,\ldots,j_{k-1}^* \right).
\end{align*}
Each set contains $(n-k)$ random variables. We prove by induction that the joint distribution of the random variables of $A_k$ is equal to those of $B_k$. Then the claim follows because $(\sum_{i=1}^k Y_{\pi_k}^{(i)})$ and $\|V_k^\top x_{j_k^*}\|_2^2$ are the maximums of $A_k$ and $B_k$, respectively.
\begin{itemize}
\item Base case : As $V_1 = x_1$, $B_1 = ( \|V_1^\top x_j\|_2^2 : w_j = 1, j \neq 1 )$ is the set of squared inner products with $x_1$ for the $n-1$ other points. Since the $n-1$ points are iid uniformly random unit vectors independent of $x_1$, the squared inner products with $x_1$ are equal in distribution to $Y_j^{(1)} = (X_{j1}^{(1)})^2$. Therefore, the joint distribution of $B_1 = ( \|V_1^\top x_j\|_2^2 : w_j = 1, j \neq 1)$ is equal to the joint distribution of $A_1 = (Y_j^{(1)} : j = 1,\ldots,n-1)$.
\item Induction : Assume that the joint distribution of $A_k$ is equal to the joint distribution of $B_k$. It is sufficient to show that given $A_k \equiv B_k$ the conditional joint distribution of $A_{k+1} = ( \sum_{i=1}^{k+1} Y^{(i)}_j : j \in [n-1], j \neq \pi_1, \ldots, \pi_k )$ is equal to the conditional joint distribution of $B_{k+1} = ( \|V_{k+1}^\top x_j\|_2^2 : w_j = 1, j \neq 1,j_1^*,\ldots,j_{k}^* )$. 
Define
$$
v_k = \frac{x_{j_k^*} - V_k V_k^\top x_{j_k^*}}{\|x_{j_k^*} - V_k V_k^\top x_{j_k^*}\|_2}.
$$
$v_k$ is the unit vector along the new orthogonal axis added on $\set{V}_k$ for $\set{V}_{k+1}$. Since we have
$$
\|V_{k+1}^\top x_j\|_2^2 = \|V_k^\top x_j\|_2^2 + (v_k^\top x_j)^2 ,\quad \forall j : w_j = 1,
$$
The two terms are independent of each other because $V_k \perp v_k$, and $x_j$ is isotropically distributed. Hence, we only need to show that $( (v_k^\top x_j)^2 : w_j = 1, j \neq 1,j_1^*,\ldots,j_{k}^* )$ is equal in distribution to $( Y^{(k+1)}_j : j \in [n-1], j \neq \pi_1, \ldots, \pi_k )$.

Since $v_k$ is a normalized vector on the subspace $\set{V}_k^\perp \cap \set{D}_1$, and $x_{j_k^*}$ is drawn iid from an isotropic distribution, $v_k$ is independent of $V_k^\top x_{j_k^*}$. Hence, the marginal distribution of $v_k$ given $\set{V}_k$ is uniform over $(\set{V}_k^\perp \cap \set{D}_1) \cap \mathbb{S}^{p-1}$. Also, $v_k$ is also independent of the points $\{x_j : w_j = 1, j \neq 1, j_1^*, \ldots, j_k^*\}$. Therefore, the random variables $(v_k^\top x_j)^2$ for $j$ with $w_j = 1, j \neq 1, j_1^*, \ldots, j_k^*$ are iid equal in distribution to $Y_j^{(k+1)}$ for any $j$.
\end{itemize}

Second, we can see that the $k$'th maximum of $\{ \sum_{i=1}^{k} Y^{(i)}_j : j \in [n-1]\}$ is equal in distribution to $P_{k,(k)}^2$. This is because each $\sum_{i=1}^{k} Y^{(i)}_j$ can be seen as the norm of the projection of a uniformly random unit vector in $\mathbb{R}^d$ onto a $k$-dimensional subspace.

Now we are ready to complete the proof. Since $\left( \sum_{i=1}^{k} Y^{(i)}_{\pi_k} \right)$ is the maximum of the $n-k$ variables of $A_k$, it is greater than or equal to the $k$'th maximum of $\left( \sum_{i=1}^{k} Y^{(i)}_j : j \in [n-1] \right)$. Therefore, $\|V_k^\top x_{j_k^*}\|_2^2$ stochastically dominates $P_{k,(k)}^2$.

The second claim is clear because $\set{V}_{\hat{k}} \subseteq \set{V}_k$, and hence the norm of the projection onto $\set{V}_k$ is always larger than the norm of the projection onto $\set{V}_{\hat{k}}$.

\subsection{Proof of Lemma \ref{lem:projbound_unitvec}a}
Let $x$ be an unit vector drawn uniformly at random from $\mathbb{S}^{d-1}$. Equivalently, $x$ can be drawn from
$$
x = \frac{w}{\|w\|_2}, \quad w \sim \mathcal{N}(0,I_{d \times d}).
$$
Define $A^\perp \in \mathbb{R}^{(d-k) \times d}$ as a matrix with orthonormal rows such that $\|w\|_2^2 = \|Aw\|_2^2 + \|A^\perp w\|_2^2$ for any $w \in \mathbb{R}^d$. We have
\begin{align}
	\Pr\left\{\|Ax\|_2^2 > \frac{k}{d}(1+\epsilon)\right\} &= \Pr\left\{\frac{\|Aw\|_2^2}{\|w\|_2^2} > \frac{k}{d}(1+\epsilon)\right\} \nonumber \\
	&= \Pr\left\{\frac{\|Aw\|_2^2}{\|Aw\|_2^2 + \|A^\perp w\|_2^2} > \frac{k}{d}(1+\epsilon)\right\} \nonumber \\
	&\ge \Pr\left\{\|Aw\|_2^2 > k(1+\epsilon), \|A^\perp w\|_2^2 < (d-k)-k\epsilon\right\} \nonumber \\
	&= \Pr\left\{\|Aw\|_2^2 > k(1+\epsilon)\right\} \cdot \Pr\left\{\|A^\perp w\|_2^2 < (d-k)-k\epsilon\right\}, \label{eqn:unitvector_projectionbound}
\end{align}
where the last equality follows from that $\|Aw\|_2$ and $\|A^\perp w\|_2$ are independent of each other because $w \sim \mathcal{N}(0, I_{d \times d})$. Note that $\|Aw\|_2^2$ and $\|A^\perp w\|_2^2$ are Chi-square random variables with degrees of freedom $k$ and $d-k$, respectively.

Now we use the following lemma.
\begin{lemma}[Chi-square upper-tail lower-bound] \label{lem:chisquare_uppertail_lowerbound}
For any $k \in \mathbb{N}$ and any $\epsilon \ge 0$, we have
$$
\Pr \{\chi^2_k \ge k (1+\epsilon)\} \ge \frac{1}{3\sqrt{k}\epsilon + 6} \exp \left(-\frac{k\epsilon}{2}\right).
$$
where $\chi^2_k$ is the chi-square random variable with $k$ degrees of freedom.
\end{lemma}

Suppose $0 \le \epsilon \le \alpha \frac{(d-k)^{\frac{1}{2}}}{k}$ for some $\alpha \in (0,1)$. It follows from Lemma \ref{lem:chisquare_uppertail_lowerbound} and the central limit theorem that
\begin{align*}
	\eqref{eqn:unitvector_projectionbound} &\ge \Pr\left\{\|Aw\|_2^2 > k(1+\epsilon)\right\} \cdot \Pr\left\{\|A^\perp w\|_2^2 - (d-k) < -\alpha(d-k)^{\frac{1}{2}} \right\} \\
	&\ge \frac{f(\alpha)}{3k\epsilon + 6} \exp \left(-\frac{k\epsilon}{2}\right)
\end{align*}
where $f(\alpha) \in (0,1)$ is some constant depending only on $\alpha$.

Then it follows that
\begin{align}
\Pr \left\{ z_{(n-m+1)}^2 < \frac{k}{d}(1+\epsilon) \right\}
&= \Pr \left\{ \exists I \subset [n], |I| = n-m+1 : z_i^2 < \frac{k}{d}(1+\epsilon), \forall i \in I \right\} \nonumber \\
&\le {n \choose m-1} \cdot \Pr \left\{ z_1^2 < \frac{k}{d}(1+\epsilon) \right\}^{n-m+1} \nonumber \\
&\le \left( \frac{ne}{m} \right)^{m} \cdot \left( 1 - \frac{f(\alpha)}{3 k\epsilon + 6} \exp\left(-\frac{k \epsilon}{2}\right)  \right)^{n-m+1} \nonumber \\
&\le \exp \left\{ m \log \frac{n e}{m} - \frac{f(\alpha) \cdot (n-m+1)}{3 k\epsilon + 6} \exp\left(-\frac{k \epsilon}{2}\right)  \right\} \label{eqn:orderstat_lowerbound}
\end{align}
where we use the facts ${n \choose m} \le \left(\frac{ne}{m}\right)^{m}$ and $1+x \le e^x, \forall x$. 

Set $C = \frac{6}{f(\alpha)}$, and choose $\epsilon$ such that
\begin{align*}
	\epsilon &= \frac{1}{k} \min \left\{ 2 \log \left( \frac{n-m+1}{C m \left( \log \frac{ne}{m\delta} \right)^2 } \right) , \alpha \sqrt{d-k} \right\}.
\end{align*}

This $\epsilon$ is valid because $0 \le \epsilon \le \alpha \frac{(d-k)^{\frac{1}{2}}}{k}$. Then we obtain
\begin{align*}
	\eqref{eqn:orderstat_lowerbound}
	&\le \exp \left\{ m \log \frac{n e}{m} - \frac{f(\alpha) \cdot (n-m+1)}{6 \log \left( \frac{n-m+1}{C m \left( \log \frac{ne}{m\delta} \right)^2 } \right) + 6} \cdot \frac{C m \left( \log \frac{ne}{m\delta} \right)^2 }{n-m+1} \right\} \\
	&= \exp \left\{ m \log \frac{n e}{m} - \frac{6 \log \frac{ne}{m\delta}}{6 \left( 1 + \log \left( \frac{f(\alpha)}{6} 
\cdot \frac{n-m+1}{m} \cdot (\log \frac{ne}{m\delta})^{-2} \right) \right)} \cdot m \log \frac{ne}{m\delta} \right\} \\
	&\le \exp \left\{ m \log \frac{n e}{m} - \frac{6 (1 + \log \frac{n}{m})}{6(1+\log \frac{f(\alpha)}{6} + \log \frac{n}{m} )} m \log \frac{ne}{m\delta} \right\} \\
	&\le \exp \left\{ m \log \frac{ne}{m} - m \log \frac{ne}{m\delta} \right\} \\
	&\le \delta^m.
\end{align*}
This completes the proof.


\subsection{Proof of Lemma \ref{lem:projbound_unitvec}b}
We use a special case of Levy's lemma for this proof.
\begin{lemma}[\cite{ledoux2005concentration}] \label{lem:measureconc}
For $x \sim \text{Unif}(\mathbb{S}^{d-1})$,
\begin{align*} 
\Pr \{ \| A x \|_2  > m\|A x\|_2 + t \} &\le \exp \left( - \frac{dt^2}{2\|A\|_2^2} \right), \\
\Pr \{ \| A x \|_2  < m\|A x\|_2 - t \} &\le \exp \left( - \frac{dt^2}{2\|A\|_2^2} \right).
\end{align*}
for any matrix $A \in \mathbb{R}^{p \times d}$ and $t > 0$. $m\|Ax\|_2$ is the median of $\|Ax\|_2$.
\end{lemma}
It follows from the lemma that
\begin{align*}
\left| E\|A x\|_2 - m\|A x\|_2 \right| &\le E\left[ \left| \|A x\|_2 - m\|A x\|_2 \right| \right] \le \int_0^\infty 2e^{-\frac{dt^2}{2\|A\|_2^2}} dt = \sqrt{\frac{2\pi}{d}} \|A\|_2.
\end{align*}
Then we have
\begin{align*}
\Pr \left\{ \|A x_i\|_2 > \sqrt{\frac{\|A\|_F^2}{d}} + \sqrt{\frac{2\pi}{d}} \|A\|_2 + t \right\}
&= \Pr \left\{ \|A x_i\|_2 > \sqrt{E\|A x_i\|_2^2} + \sqrt{\frac{2\pi}{d}} \|A\|_2 + t \right\} \\
&\le \Pr \left\{ \|A x_i\|_2 > E\|A x_i\|_2 + \sqrt{\frac{2\pi}{d}} \|A\|_2 + t \right\} \\
&\le \Pr \left\{ \|A x_i\|_2 > m\|A x_i\|_2 + t \right\} \\
&\le \exp \left( -\frac{dt^2}{2 \|A\|_2^2} \right).
\end{align*}
It follows that
\begin{align*}
&\Pr \left\{ z_{(n-m+1)} > \sqrt{\frac{\|A\|_F^2}{d}} + \sqrt{\frac{2\pi}{d}} \|A\|_2 + t \right\} \\
&\le \Pr \left\{ \exists I \subset [n], |I| = m : \|Ax_i\|_2 > \sqrt{\frac{\|A\|_F^2}{d}} + \sqrt{\frac{2\pi}{d}} \|A\|_2 + t, \forall i \in I \right\} \\
&\le {n \choose m} \cdot \Pr \left\{ \|Ax_1\|_2 > \sqrt{\frac{\|A\|_F^2}{d}} + \sqrt{\frac{2\pi}{d}} \|A\|_2 + t \right\}^{m} \\
&\le \left( \frac{n e}{m} \right)^{m} \cdot \exp \left( - \frac{mdt^2}{2 \|A\|_2^2} \right) \\
&= \exp \left\{ m \log \frac{ne}{m} -\frac{md t^2}{2 \|A\|_2^2} \right\}.
\end{align*}
Replacing $t$ with $\sqrt{\frac{2\|A\|_2^2}{d} \log \frac{ne}{m \delta}}$, we obtain the desired result.

\subsection{Proof of Lemma \ref{lem:subspace_concentration}a}
Let $A = U \Sigma V^\top$ be the singular value decomposition of $A$. Then we have
\begin{align*}
E[\|AX\|_F^2] = E[\|U \Sigma V^\top X\|_F^2] = E[\|\Sigma X\|_F^2] = \sum_{i=1}^{\min(p,d)} \sigma_i^2 \cdot \left( \sum_{j = 1}^k E[X_{ij}^2] \right) = \sum_{i=1}^{\min(p,d)} \sigma_i^2 \cdot \frac{k}{d} = \frac{k}{d} \|A\|_F^2.
\end{align*}
where the second last equality follows from that $X_{ij}$ is a coordinate of a uniformly random unit vector, and thus
\begin{align*}
E[X_{ij}^2] = \frac{1}{d} ,\quad \forall i,j.
\end{align*}

\subsection{Proof of Lemma \ref{lem:subspace_concentration}b}
Consider the Stiefel manifold $V_k(\mathbb{R}^d)$ equipped with the Euclidean metric. We see that $X$ is drawn from $V_k(\mathbb{R}^d)$ with the normalized Harr probability measure. We have
$$
\|AX\|_F - \|AY\|_F \le \|AX - AY\|_F = \|A(X-Y)\|_F \le \|A\|_2 \|X-Y\|_F
$$
for any $X, Y \in \mathbb{R}^{d \times k}$. Since $\|A\|_2 \le 1$, $\|A X\|_F$ is a 1-Lipschitz function of $X$. Then it follows from \cite{milman1986asymptotic,ledoux2005concentration} that
\begin{align*}
\Pr \{ \|A X \|_F > m\|A X\|_F + t \} &\le e^{-\frac{(d-1)t^2}{8}},
\end{align*}
where $m\|A X\|_F$ is the median of $\|A X\|_F$. Also, we have
\begin{align*}
\Pr \{ | \|A X \|_F - m\|A X\|_F| > t \} &\le 2e^{-\frac{(d-1)t^2}{8}},
\end{align*}
and then it follows that
\begin{align*}
\left| E\|A X\|_F - m\|A X\|_F \right| &\le E\left[ \left| \|A X\|_F - m\|A X\|_F \right| \right] \le \int_0^\infty 2e^{-\frac{(d-1)t^2}{8}} dt = \sqrt{\frac{8\pi}{d-1}}.
\end{align*}
It follows from Jensen's inequality and Lemma \ref{lem:subspace_concentration}a that
\begin{align*}
E\|A X\|_F \le \sqrt{E\|A X\|_F^2} = \sqrt{\frac{k}{d}}\|A\|_F
\end{align*}
Putting the above inequalities together using the triangle inequality, we obtain the desired result.

\subsection{Proof of Lemma \ref{lem:chisquare_uppertail_lowerbound}}
For $k \ge 2$, it follows from \cite[Proposition 3.1]{inglot2010chisquare} that
\begin{align*}
\Pr \{\chi^2_k \ge k (1+\epsilon)\} &\ge \frac{1-e^{-2}}{2} \frac{k(1+\epsilon)}{k\epsilon + 2\sqrt{k}} \exp \left( -\frac{1}{2} (k\epsilon - (k-2)\log(1+\epsilon) + \log k) \right) \\
&\ge \frac{1}{3\sqrt{k}\epsilon + 6} \exp \left( -\frac{k}{2} (\epsilon - \log(1+\epsilon)) \right) \\
&\ge \frac{1}{3\sqrt{k}\epsilon+6} \exp \left( -\frac{k \epsilon}{2} \right).
\end{align*}
For $k = 1$, we can see numerically that the inequality holds.

\end{document}